\pdfoutput=1

\documentclass[11pt]{article}
\usepackage{booktabs}
\usepackage{multirow}

\usepackage[final]{acl}

\usepackage{algorithm}
\usepackage{times}
\usepackage{latexsym}
\usepackage{algorithm}
\usepackage{algorithmicx}
\usepackage{algpseudocode}
\usepackage{amsmath}

\newcommand{\kwin}{\textbf{Input:} }
\newcommand{\kwout}{\textbf{Output:} }

\usepackage[T1]{fontenc}

\usepackage[utf8]{inputenc}

\usepackage{microtype}

\usepackage{inconsolata}

\usepackage{graphicx}

\title{Knockout LLM Assessment: Using Large Language Models for Evaluations through Iterative Pairwise Comparisons}

\author{Isik Baran Sandan, Tu Anh Dinh, Jan Niehues \\Karlsruhe Institute of Technology
  }

\begin{document}
\maketitle

\begin{abstract}
\label{abstract}

Large Language Models (LLMs) have shown to be effective evaluators across various domains such as machine translations or the scientific domain. Current LLM-as-a-Judge approaches rely mostly on individual assessments or a single round of pairwise assessments, preventing the judge LLM from developing a global ranking perspective.
To address this, we present Knockout Assessment, an LLM-as-a-Judge method using a knockout tournament system with iterative pairwise comparisons. 
Experiments across three LLMs on two datasets show that knockout assessment improves scoring accuracy, increasing Pearson correlation with expert evaluations by 0.07 on average for university-level exam scoring and machine translation evaluations, aligning LLM assessments more closely with human scoring.

\end{abstract}

\section{Introduction}
\label{introduction}

Across various domains, and especially for scientific research, accurate and consistent evaluations are very crucial for informed decision-making. However, the inherent scale and subjectivity make this task very challenging and time-consuming. In recent years, the methodology of “LLM-as-a-Judge” \cite{mtbench} has emerged to tackle this challenge, where instead of humans, Large Language Models (LLMs) take the role of the expert to evaluate complex tasks. Using LLMs as evaluators allows us to mimic the abilities of human experts, making evaluations cost-effective and scalable.

Although many approaches to LLM-as-a-Judge exist, the most common is individual assessment, in which the evaluation prompt consists of only the question and the corresponding answer, which is to be evaluated \cite{individualAssessment1}. While this approach has already shown to yield good evaluation results next to providing scalability \cite{individualAssessment1, dinh2024sciexbenchmarkinglargelanguage}, it does not consider the relative strength of answers in a set to a given question. The more recent approach of pairwise assessment tries to address this issue by providing two responses to the judge LLM each time, however, it still fails to account for a global ranking perspective, as pairwise comparisons do not analyze how all responses compare to each other in the broader sense.

In this paper we present an LLM-as-a-judge method called \textbf{Knockout Assessment} to address this challenge, which can be seen as a variation of the tournament system used by \citet{mtbench}, differing in that it makes use of iterative pairwise comparisons. Instead of isolating responses individually or in pairs for evaluation, Knockout Assessment focuses on an iterative process where responses are compared against one another multiple times in a tournament manner. In each round, stronger responses advance to compete against each other in later rounds, allowing us to refine the scores progressively throughout the tournament.  This approach allows the judge LLMs to develop a global perspective on responses without requiring all replies to be included in a single prompt, which would otherwise result in an impractically long context length.

To summarize, our contributions are as follows: 
\begin{itemize}
    \item Knockout Assessment, an LLM-as-a-Judge methodology which makes use of iterative pairwise comparisons in a knockout tournament system for more accurate evaluations
    \item Analysis of Knockout Assessment's performance compared to individual assessment's and naive pairwise assessment's performance on two different datasets concerning scientific evaluation and machine translation evaluation.
\end{itemize}

\section{Related Work}

\paragraph{Individual Assessment}  One approach to LLM-as-a-Judge is individual assessment, where the judge LLM is provided with a prompt or a question, the corresponding answer, the scoring criteria and is asked to provide an evaluation such as a grade. Various studies have used this method for tasks such as evaluating story generation, scoring quality of different texts according to different criteria, or grading university-level exams \cite{individualAssessment1, individualAssessment2, dinh2024sciexbenchmarkinglargelanguage}. 

\paragraph{Pairwise Assessment}  Another more recent LLM-as-a-Judge approach is pairwise prompting, in which the LLM judge is provided with two responses to the prompt instead of one. The judge LLM is then asked to evaluate both responses. This has shown to be an effective LLM-as-a-Judge method for ranking documents, as it gives the judge LLM direct comparison points while making evaluations \cite{pairwisePrompting}. However, this method still does not make the Judge LLM develop a global ranking perspective.

\paragraph{Chatbot Arena Approach}   
\citet{mtbench} made use of an ELO system in which all possible answer pairs are evaluated against one another. This approach is thus able to make implicit use of a global view over the dataset while assigning scores. However, pairing all possible answers results in a computational time of $O(N^2)$.

\paragraph{Sorting Based Approaches}  To address this inefficiency, \citet{qin2024largelanguagemodelseffective} introduced two new methods. First approach uses Heapsort with pairwise comparisons to sort out the possible answers ($O(NlogN)$). Second is a sliding window approach, making use of individual passes in the Bubble Sort algorithm for a constant number $K$ times ($O(N)$).

\section{Knockout Assessment}

We propose using multiple iterative pairwise comparisions instead of individual assessment with Knockout Assessment. In each pairwise assessment, one pairwise ranking prompt similar to the comparative prompt introduced by \citet{ pairwisePrompting} is used. In each prompt, one question and two answers to that question are provided to the judge LLM, which is asked to evaluate both of those answers. We call this a “question-level-match”. The exact prompts we used for our experiments can be found in Appendix \ref{sec:appendix:Prompts}.

From the response generated by the judge LLM, the score each individual answer got is parsed and saved to the list of scores for that answer, which keeps track of all the scores an answer got throughout all its question-level-matches. The answer which got the higher score advances to the next round to be matched up against another answer. 

The order of texts in pairwise rankings has shown to be an influential factor in the LLMs decision making \cite{resnik2024largelanguagemodelsbiased}, thus we also collected the results with using a debiasing methodology similar to the one introduced by \citet{pairwisePrompting}, averaging scores from both possible orderings of each answer pair.  Debiasing thus results in double the compute-time compared to a regular question-level-match.

The main methodology behind our appproach is a knockout tournament system that iteratively uses the question-level matches. In each tournament round, the $N$ available answers to a question are randomly assgined to pairs. Each pair then enters a question-level match, and the higher scoring response advances to the next round. In the case when $N$ is odd, one answer directly advances to the next round. This continues until we reach a tournament round with a single response. 

\begin{figure}[htbp]
  \includegraphics[width=\columnwidth]{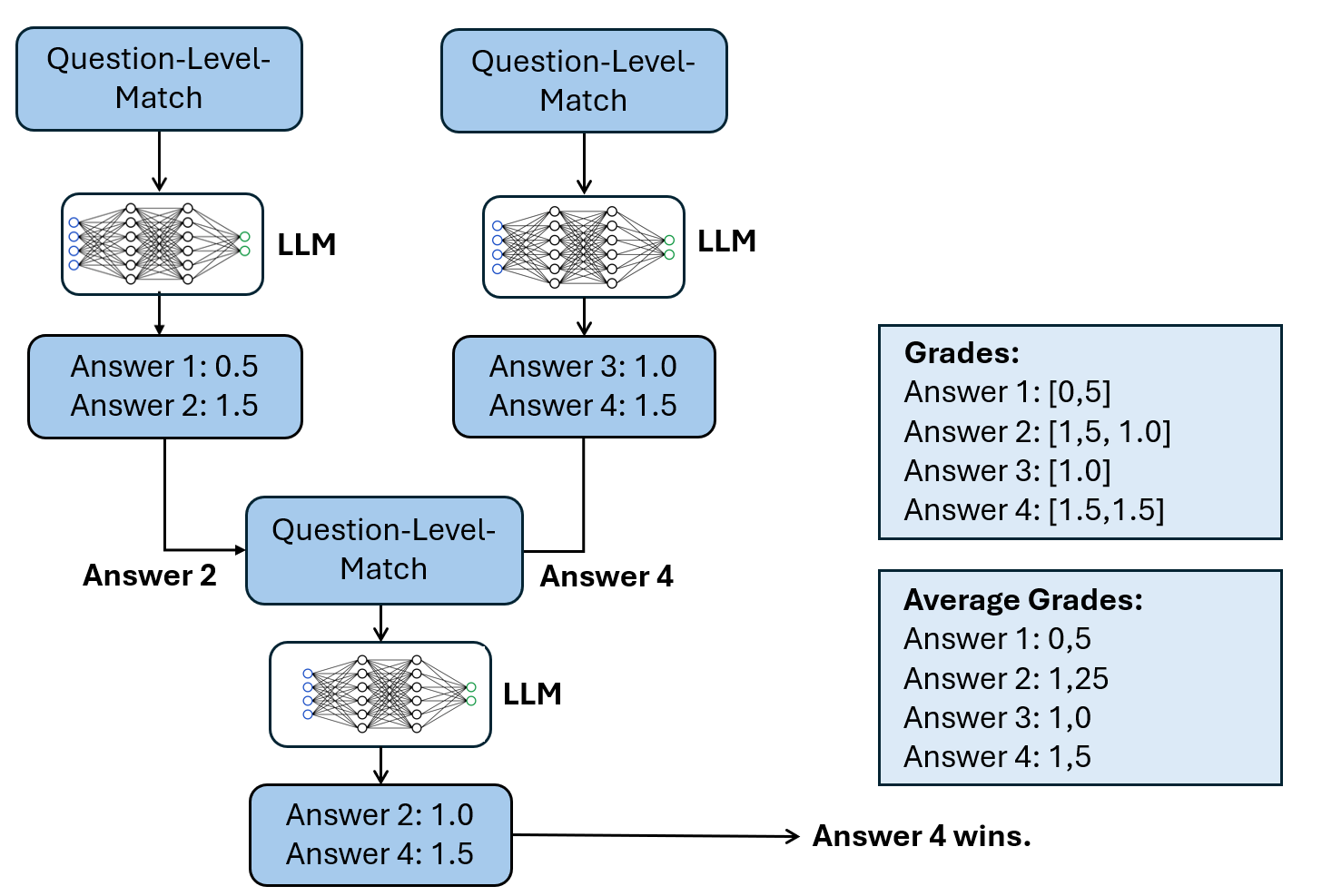}
  \caption{An example Knockout-Tournament with 4 answers for a question.}
  \label{fig:exampleTournament}
\end{figure}

Once the tournament ends, the final evaluation score for each answer is computed as the average of all the scores it received throughout the tournament. An example tournament with N = 4 answers is depicted in Figure \ref{fig:exampleTournament}. The full algorithm is given in \ref{sec:Appendix:algo}.

\section{Experiments}


\paragraph{Datasets} Overall, the datasets we used include the task input, machine outputs, and human-assgined scores on the machine output.  The first dataset is SciEx \cite{dinh2024sciexbenchmarkinglargelanguage}, which consists of university exam questions, LLM answers, human grades and LLM grades. Each question is labeled by difficulty and language.
SciEx contains 1120 question-answer pairs in total. The second dataset is taken from the WMT Metrics Shared Task \cite{kocmi-etal-2024-findings}, which includes a list of source sentences and machine translations, accompanied by a human evaluation on a scale of 0 to 100. We filtered the dataset by languages supported by the Llama models (English, German, French, Italian, Portuguese, Hindi, Spanish, and Thai). We also remove identical translations of the same source sentence, as sometimes the human evaluations differed for the same translation, causing inconsistencies. The final processed WMT dataset includes 2100 source-translation pairs.

\paragraph{Baselines} We compare our approach to 2 baselines: (1) individual assessment, where each answer is scored individually and (2) naive pairwise assessment, where answers are paired up and scored only one time, without multiple knockout rounds. 

\paragraph{Evaluation Metrics}  We use Pearson correlation as our primary evaluation metric, which measures the linear relationship between the scores provided by our method and the human-provided scores.
Additionally, we use pairwise ranking accuracy \cite{kocmi-etal-2021-ship}, which indicates how often the LLM judges select the correct winning answer given a pair of answers.
The results for pairwise ranking accuracy can be found in Appendix \ref{sec:Appendix:pra} since they mirror our Pearson correlation findings.

\paragraph{Models}  The models we used as judges for our experiments are Meta’s Llama 3.2 1B parameter model, Llama 3.2 3B parameter model and Llama 3.1 70B parameter model. All the model checkpoints for our experiments were obtained from the HuggingFace model hub. For consistency with similar work \cite{dinh2024sciexbenchmarkinglargelanguage} the temperature parameter of the models was set to 0.1 for our experiments.

\paragraph{Hardware}  The experiments on the 70B parameter model were conducted on 4 NVIDIA Tesla V100 GPUs with 32GB VRAM each. The experiments for the smaller models were conducted on 1 NVIDIA Tesla V100 GPU with 32GB VRAM.

\subsection{General Results}

\begin{table*}
    \centering
    \small
    \begin{tabular}{c ccc ccc ccc c}
        \toprule
        Method & \multicolumn{3}{c}{SciEx Question-Level} & \multicolumn{3}{c}{SciEx Exam-Level} & \multicolumn{3}{c}{WMT Dataset} & Overall \\
        \cmidrule(lr){2-4} \cmidrule(lr){5-7} \cmidrule(lr){8-10}
        & 3.2 1B & 3.2 3B & 3.1 70B & 3.2 1B & 3.2 3B & 3.1 70B & 3.2 1B & 3.2 3B & 3.1 70B & . \\
        \midrule
        Ind. Assessment & 0.400 & 0.365 & 0.615 & 0.504 & 0.465 & 0.667 & 0.050 & 0.187 & \textbf{0.397} & 0.405 \\
        KO. Assessment & 0.434 & 0.541 & 0.622 & 0.627 & \textbf{0.555} & 0.652 & \textbf{0.113} & 0.207 & 0.222 & 0.441 \\
        (No Debiasing) \\
        KO. Assessment & \textbf{0.443} & \textbf{0.558} & \textbf{0.648} & \textbf{0.672} & 0.540 & \textbf{0.697} & 0.087 & \textbf{0.259} & 0.268 & \textbf{0.475} \\
        (Debiased)\\
        \bottomrule
    \end{tabular}
    \caption{LLM scores' Pearson correlation to expert scores for different datasets, subdivided by models}
    \label{tab:general_table}
\end{table*}


From Table \ref{tab:general_table}, it can be seen that our knockout assessment method improves performance on all datasets and model judges compared to individual assessment. Generally, the scores sampled using debiasing have improved performance even further.

One failing case is on the WMT subset, where the best performing assessment method is to use individual assessment with Llama 3.1 70B. In this case, knockout assessment has
decreased performance for the largest model. However, smaller models still saw an increase in performance with knockout assessment compared to individual assessment. Overall, we see an average increase of 0.07 Pearson correlation over individual assessment across all our experiments, when debiased knockout assessment is used.

One possible reason for this is that, Knockout Assessment is more useful when the evaluation task is more complex. Therefore, it consistently helps improving the assessment performance on the SciEx dataset, which contains difficult university-level scientific question-answer pairs. On the other hand, evaluating the machine translation task is more simple, thus a large model like Llama 3.1 70B can have good performance with just individual assessment. In this case, introducing other answers with Knochout Tournament could introduce noise, thus lower the performance.

\subsection{Comparison Against Naive Pairwise Assessment}

\begin{table*}
\centering
\small
\begin{tabular}{llcccccccc}
\toprule
Dataset & Elimination & \multicolumn{8}{c}{Knockout Assessment} \\
\cmidrule(lr){3-10}
& & \multicolumn{2}{c}{\rotatebox{0}{1B}} & \multicolumn{2}{c}{\rotatebox{0}{3B}} & \multicolumn{2}{c}{\rotatebox{0}{70B}} & \multicolumn{2}{c}{\textbf{Overall}} \\
& & \rotatebox{45}{Biased} & \rotatebox{45}{Debiased} & \rotatebox{45}{Biased} & \rotatebox{45}{Debiased} & \rotatebox{45}{Biased} & \rotatebox{45}{Debiased} & \rotatebox{45}{\textbf{Biased}} & \rotatebox{45}{\textbf{Debiased}}\\
\midrule
\multirow{3}{*}{SciEx} & First Round & 0.3737 & 0.3223 & 0.5400 & 0.5400 & 0.5264 & 0.5801 & 0.4800 & 0.4808\\
& Later Rounds & 0.4816 & 0.4428 & 0.5393 & 0.5692 & 0.6602 & 0.6782 & 0.5604 & 0.5634\\
& Difference & +0.1079 & +0.1205 & -0.0007 & +0.0292 & +0.1338 & +0.0981 & +0.0804 & +0.0826\\
\midrule
\multirow{3}{*}{WMT} & First Round & 0.1245 & 0.0836 & 0.2222 & 0.2917 & 0.2688 & 0.3173 & 0.2052 & 0.2309\\
& Later Rounds & 0.0777 & 0.0875 & 0.1603 & 0.2013 & 0.1150 & 0.1581 & 0.1177 & 0.1490\\
& Difference & -0.0468 & +0.0039 & -0.0619 & -0.0904 & -0.1538 & -0.1592 & -0.0875 & -0.0819\\
\bottomrule
\end{tabular}
\caption{Comparison of LLM Grader's performance on SciEx and WMT datasets for answers graded once versus multiple times.}
\label{tab:combinedElimination}
\end{table*}

In this section, we check whether knockout assessment results in any additional performance increase compared to regular pairwise comparisons with only one round. We report the Pearson correlations of the sets of scores the answers got, based on their round of elimination.

For both datasets, the answers/translations which got eliminated on the first round of the knockout tournament, got only one pairwise comparison, compared to the multiple pairwise comparisons the answers/translations which advanced to later rounds got. As can be seen in Table \ref{tab:combinedElimination}, for SciEx, the answers which got eliminated in later rounds have an overall higher Pearson correlation in the grades they got, across the three models we used. This shows that more pairwise comparisons result in more accurate grades from the judge LLM.

However, as can be seen in Table \ref{tab:combinedElimination}, the responses which advanced further in the tournament showed lower alignment with human experts in the WMT dataset. This suggests that the iterative comparisons do not increase the scoring accuracy for the task of machine translation, but rather introduce noises.

\subsection{Effect of Difficulty Levels}
\begin{figure}[htbp]
    \centering
    \includegraphics[width=\columnwidth]{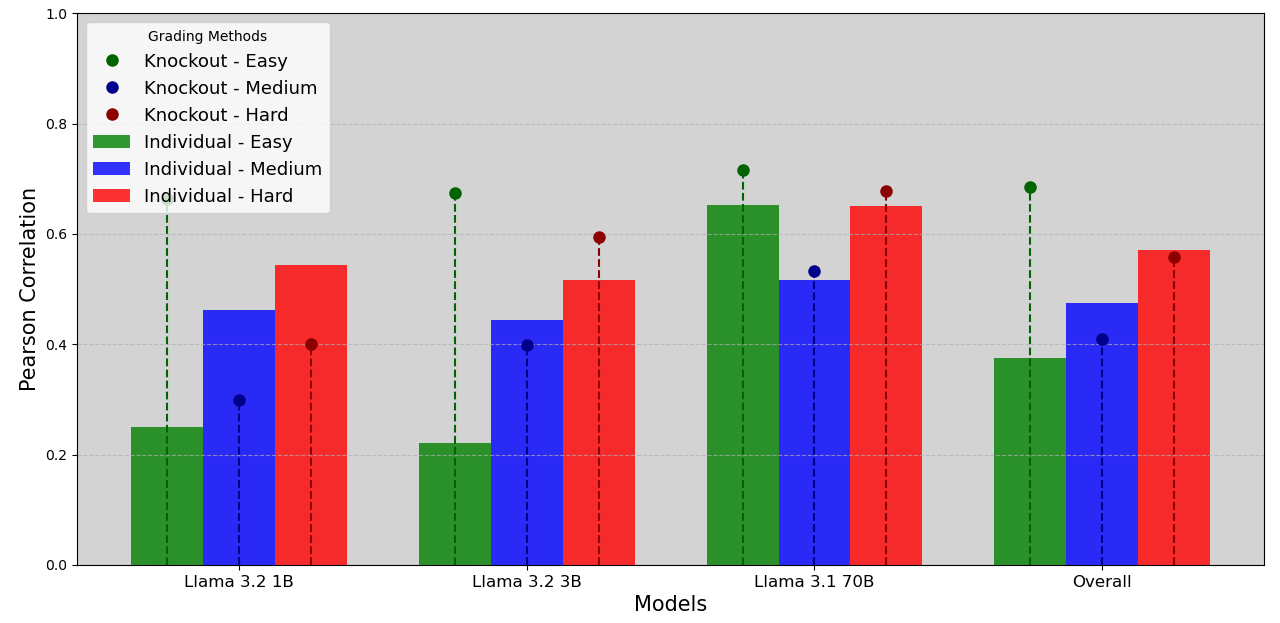} 
    \caption{Performance by Model and Difficulty Level: Knockout (debiased) vs. Individual Assessment}
    \label{fig:Difficulty}
\end{figure}

We investigate how the difficulty level of the task effect the performance of our assessment method. We use the question-level difficulty labels from SciEx, and report the performance splitted by the labels of "Easy", "Medium" and "Hard".
The results are shown in Figure \ref{fig:Difficulty}. As can be seen, for individual assessment, the models perform better on scoring answers of difficult questions than easier questions, which is rather counter-intuitive. However, this aligns with the finding by \citet{dinh2024sciexbenchmarkinglargelanguage} that LLMs can perform worse on easy questions, since they may lack specific course knowledge compared to the students.

With our Knockout Assessment method, the performance of the judges on easy questions significantly increases. This shows that, by including information of multiple candidate answers, we give more global view to the LLMs, thus help them to better provide assessment scores, even when their own internal knowledge for the question is lacking.

\section{Conclusion}
We address a key limitation of many existing LLM-as-a-Judge methods: not having a global view over the responses while evaluating them. To address this, we proposed an alternative LLM-as-a-Judge method called knockout assessment and tested it with three different LLMs on two different datasets. 
Knockout Assessment improves Pearson correlation to human evaluations by 0.07 over individual assessment on average. The performance increase was more significant in scientific evaluation compared to machine translation evaluation, especially for the larger LLM. 
Furthermore, for scientific evaluation, the responses which progressed further in the knockout assessment process had 0.08 better accuracy compared to the responses which got eliminated on the first round, which indicates that knockout assessment results in a performance increase from regular pairwise assessments.

\section*{Acknowledgments}
This work was supported by the Helmholtz Programme-oriented Funding, with project number 46.24.01, project name AI for Language Technologies. 
We acknowledge the HoreKa supercomputer funded by the Ministry of Science, Research and the Arts Baden-Wurttemberg and by the Federal Ministry of Education and Research.

\bibliography{custom}
\newpage

\appendix

\section{Appendix}
\label{sec:appendix}

\subsection{Algorithm for Knockout Assessment}
\label{sec:Appendix:algo}

\begin{algorithm}[H]
\caption{Knockout Assessment}
\kwin{prompt $P$, Set of responses $\mathcal{R}$, scoring function $G$ (evaluates response quality)}
\kwout{Final Champion Response, All Scores, Final Average Scores}
\begin{algorithmic}[1]
    
    \While{$|\mathcal{R}| > 1$}
        \State Form \textbf{consecutive pairs} of responses from $\mathcal{R}$
        \For{$k = 1, 3, 5, \dots, |\mathcal{R}| - 1$} \Comment{Match responses in consecutive pairs}
            \State $(\text{Score1}, \text{Score2}) \gets \texttt{Question\_Level\_Match}(P, G, r_k, r_{k+1})$
            \State Update \texttt{All Scores} with \text{Score1} and \text{Score2}
            \If{$\text{Score1} > \text{Score2}$}
                \State Advance $r_k$ to the next round
            \Else
                \State Advance $r_{k+1}$ to the next round
            \EndIf
        \EndFor
        \If{$|\mathcal{R}|$ is odd}
            \State Advance the unmatched response to the next round
        \EndIf
    \EndWhile
    \State Compute \texttt{Final Average Scores} for each response across rounds
    \State \textbf{answer}: Final Champion $r^*$ (last remaining response), \texttt{Final Average Scores}, \texttt{All Scores}
    
\end{algorithmic}
\label{algo1}
\end{algorithm}

\subsection{Prompts}
\label{sec:appendix:Prompts}

This section includes all the prompts we used to generate scores for our experiments.

\textbf{Individual Grading Prompt in English:}  You are a university professor exam grader. Grade the following answer on a scale of 0 to <maxpoints> (allowing half points) based on its correctness and relevancy given the following question.

The Question: <Question>

The Answer: <Answer 1>

Give your grade in the format: Explanation: [explanation]  Score: [score]/<maxpoints>

\textbf{Pairwise Grading Prompt in English:}  You are a university professor exam grader. Grader the following answers on a scale of 0 to <maxpoints> (allowing half points) based on how well they answer the question.

The Question: <Question>

Answer 1: <Answer 1>

Answer 2: <Answer 2>

Grade the 2 answers on a scale of 0 to <maxpoints> (Half points such as 0.5 or 1.5 are allowed.) in the format: Explanation: [explanation]  Answer 1: X/<maxpoints>  Answer 2: Y/<maxpoints>

\textbf{Individual Grading Prompt in German:}  Sie sind ein Universitätsprofessor. Bewerten Sie die folgende Antwort auf die unten stehende Frage. Geben Sie eine Punktzahl von 0 bis {maxpoints} basierend auf Korrektheit und Relevanz an.

Die Frage: <question>

Die Antwort: <Answer 1>

Bewerten Sie die Antworte auf einer Skala von 0 bis {maxpoints} (halbe Punkte wie 0,5 oder 1,5 sind erlaubt) im Format: Begründung:[begründung]   Punktzahl: X/<maxpoints>

\textbf{Pairwise Grading Prompt in German:}  Sie sind ein Universitätsprofessor und bewerten Prüfungsantworten. Bewerten Sie die folgenden Antworten auf einer Skala von 0 bis {maxpoints} (halbe Punkte sind erlaubt) basierend darauf, wie gut sie die Frage beantworten.

Die Frage: {question}

Antwort 1: {answer1} 

Antwort 2: {answer2}

Bewerten Sie die beiden Antworten auf einer Skala von 0 bis {maxpoints} (halbe Punkte wie 0,5 oder 1,5 sind erlaubt) im Format: Begründung: [begründung]   Antwort 1: X/{maxpoints} Antwort 2: Y/{maxpoints}.

\textbf{Individual Scoring Prompt for MT:}  You are a translation evaluator. Evaluate the quality of the translation provided. Give a score from 0 to 100 based on clarity, accuracy and grammar.

Source: <source>

Translation: <tgt>

Output only: : Explanation: [explanation]  Score: [score]/100

\textbf{Pairwise Scoring Prompt for MT:}  You are a translation evaluator. Your task is to evaluate the quality of two translations for a given source sentence. You will provide a score from 0 to 100, based solely on clarity, accuracy and grammar of the translations.

Source: <source>

Translation 1: <tgt1>

Translation 2: <tgt2>

Output only:  Explanation: [explanation]    Translation 1: [score]/100   Translation 2: [score]/100

\subsection{Pairwise Ranking Accuracy Results}
\label{sec:Appendix:pra}

The performance of each evaluation method with pairwise ranking accuracy as an evaluation metric can be seen in Table \ref{tab:pra}, divided by the grading LLM. 

\begin{table}[h]
    \centering
    \small
    \setlength{\tabcolsep}{3pt} 
    \begin{tabular}{c @{}cccc@{}}
        \hline
        Method & \multicolumn{3}{c}{SciEx Exam Level} & Overall \\
        \cline{2-4}
        & 3.2 1B & 3.2 3B & 3.1 70B & \\
        \hline
        Ind. Assessment & 0.529 & 0.533 & 0.695 & 0.586 \\
        KO. Assessment & 0.557 & 0.543 & 0.676 & 0.592 \\
        Debiased KO. Assessment  & \textbf{0.610}  & \textbf{0.591} & \textbf{0.767} & \textbf{0.656} \\
        \hline
    \end{tabular}
    \caption{LLM scores' pairwise ranking accuracy to expert scores for different methods}
    \label{tab:pra}
\end{table}

\subsection{Influential Factors for SciEx}
\label{sec:Appendix:influentialFactors}
Our findings for the impact of knockout assessment on different examinees and different languages can be seen in Tables \ref{tab:gradersChange} and \ref{tab:languagbased}

\begin{table}[h!]
\centering
\small
\begin{tabular}{l ccc c}
\toprule
Examinee & \multicolumn{3}{c}{LLama Models} & \textbf{Overall} \\
\cmidrule(lr){2-4}
& 1B & 3B & 70B & \\
\midrule
Llava   & +0.0150 & -0.0040  & +0.0724  & \textbf{+0.0283}  \\
Mistral & +0.2452 & +0.1043  & -0.0247  & \textbf{+0.1082}  \\
Mixtral & -0.1247 & +0.0407  & -0.0776  & \textbf{-0.0539}  \\
Qwen    & -0.1263 & +0.0184  & -0.0106  & \textbf{-0.0395}  \\
Claude  & -0.2975 & +0.0760  & -0.0018  & \textbf{-0.0744}  \\
GPT-3.5 & -0.0310 & +0.1943  & +0.0228  & \textbf{+0.0620}  \\
GPT-4V  & +0.1903 & +0.3529  & +0.0046  & \textbf{+0.1826}  \\
\bottomrule
\end{tabular}
\caption{Performance difference with knockout assessment vs. individual assessment, by examinee and model.}
\label{tab:gradersChange}
\end{table}

\begin{table*}
\centering
\small
\begin{tabular}{lccccccccc}
\toprule
Language & \multicolumn{3}{c}{Individual} & \multicolumn{6}{c}{Knockout} \\
\cmidrule(lr){2-4} \cmidrule(lr){5-10}
 & 1B & 3B & 70B & \multicolumn{2}{c}{1B} & \multicolumn{2}{c}{3B} & \multicolumn{2}{c}{70B} \\
 & & & & Biased & Debiased & Biased & Debiased & Biased & Debiased\\
\midrule
English & 0.2348 & 0.176 & 0.6759 & 0.5695 & 0.5691 & 0.6365 & 0.6892 & 0.6429 & 0.6952 \\
German  & 0.5474 & 0.5451 & 0.6263 & 0.3706 & 0.3824 & 0.5456 & 0.5628 & 0.6497 & 0.6790 \\
\bottomrule
\end{tabular}
\caption{Pearson correlations for LLM graders' performance across languages (English and German), for individual and knockout assessment.}
\label{tab:languagbased}
\end{table*}

\end{document}